\documentclass[conference]{IEEEtran}

\IEEEoverridecommandlockouts
\usepackage{cite}
\usepackage{amsmath,amssymb,amsfonts, amsthm}
\usepackage{algorithmic, algorithm, }
\usepackage{threeparttable}
\usepackage{textcomp}
\usepackage{xcolor}
\usepackage{subfigure}
\usepackage{booktabs}
\usepackage{array}
\usepackage{multirow}
\usepackage{setspace}
\usepackage{graphicx}
\usepackage{caption}
\usepackage{subcaption}

\def\BibTeX{{\rm B\kern-.05em{\sc i\kern-.025em b}\kern-.08em
    T\kern-.1667em\lower.7ex\hbox{E}\kern-.125emX}}
\begin{document}

\title{CUER: Corrected Uniform Experience Replay for Off-Policy Continuous Deep Reinforcement Learning Algorithms}

\author{\IEEEauthorblockN{Arda Sarp Yenicesu\textsuperscript{1*}, Furkan B. Mutlu\textsuperscript{2*}, Suleyman S. Kozat\textsuperscript{2}$\dagger$, Ozgur S. Oguz\textsuperscript{1}}
\IEEEauthorblockA{\textsuperscript{1}\textit{Computer Engineering Department, Bilkent University, Ankara, Turkey}}
\IEEEauthorblockA{\textsuperscript{2}\textit{Electrical and Electronics Engineering Department, Bilkent University, Ankara, Turkey}}
\IEEEauthorblockA{ \{sarp.yenicesu, ozgur.oguz\}@bilkent.edu.tr}
\IEEEauthorblockA{ \{burak.mutlu; kozat\}@ee.bilkent.edu.tr}
\IEEEauthorblockA{$\dagger$IEEE Senior Member}
\thanks{*The first two authors contributed equally.}
}

\maketitle 
\begin{abstract}
The utilization of the experience replay mechanism enables agents to effectively leverage their experiences on several occasions. In previous studies, the sampling probability of the transitions was modified based on their relative significance. The process of reassigning sample probabilities for every transition in the replay buffer after each iteration is considered extremely inefficient. Hence, in order to enhance computing efficiency, experience replay prioritization algorithms reassess the importance of a transition as it is sampled. However, the relative importance of the transitions undergoes dynamic adjustments when the agent's policy and value function are iteratively updated. Furthermore, experience replay is a mechanism that retains the transitions generated by the agent's past policies, which could potentially diverge significantly from the agent's most recent policy. An increased deviation from the agent's most recent policy results in a greater frequency of off-policy updates, which has a negative impact on the agent's performance. In this paper, we develop a novel algorithm, Corrected Uniform Experience Replay (CUER), which stochastically samples the stored experience while considering the fairness among all other experiences without ignoring the dynamic nature of the transition importance by making sampled state distribution more on-policy. CUER provides promising improvements for off-policy continuous control algorithms in terms of sample efficiency, final performance, and stability of the policy during the training.

\end{abstract}


\section{Introduction}
Deep Reinforcement Learning algorithms have demonstrated significant achievements in problems that require the execution of sequential decision-making processes in various domains, such as ATARI games \cite{mnih2013playing}, continuous control tasks \cite{lillicrap2019continuous}, board games \cite{go}, and real-time strategy games \cite{2019Natur.575..350V}, where in all cases they demonstrated exceptional performance beyond human capabilities. The utilization of function approximation, such as neural networks, in reinforcement learning allows the agent to acquire a parameterized policy and reach a state of convergence towards an almost optimal policy, without the need to visit every state-action pair \cite{Sutton1998}. However, if the inputs are streamed through the agent's current experiences, this will create a temporally correlated input pattern and the process of providing temporally correlated inputs to the policy network goes against the assumption of independent and identically distributed (i.i.d.) data, which is essential to stochastic gradient-based optimization methods \cite{doi:10.1137/20M1316640}. 

Experience replay is a fundamental concept in reinforcement learning that tackles the given problem by storing past experiences in a buffer and reusing them for stability, decorrelation, data, and sample efficiency \cite{lin_er}. This concept consists of storing and later sampling past experiences to improve an agent, and sampling typically refers to uniform sampling among the experiences stored until that time. The utilization of uniform sampling results in a sampled batch of decorrelated data that exhibits similar characteristics to the independently and identically distributed (i.i.d.) scenario. This similarity improves the effectiveness of the learning process, as demonstrated in numerous studies \cite{mnih2013playing,mnih2015human,andrychowicz2017hindsight,vanhasselt2015deep,wang2016sample}.

% Nevertheless, it is essential to note that the uniform sampling exhibits an apparent contradiction. Specifically, the use of uniform sampling, which aims to ensure fairness in each sampling step, fails to account for the expansion of the replay buffer during each transition step. Consequently, this expansion introduces a bias towards earlier transitions, as they are sampled more frequently compared to more recent transitions which shifts the state distribution to a more off-policy distribution. As a result, this shift in the state distribution increases the likelihood of the occurrence of soft divergence while decreasing the overall performance of the agent \cite{vanhasselt2018deep}.

However, it is essential to note that uniform sampling introduces a subtle bias. While uniform sampling seems fair at first glance, it unintentionally favors older transitions as the replay buffer grows with each new experience.  This is because these older transitions have a higher chance of being selected simply by being present for a longer time. This bias towards older data skews the distribution of states the agent learns from, making it more "off-policy." In reinforcement learning, "off-policy" refers to situations where the agent is learning from experiences collected under a different policy than the one it's currently using. This mismatch between past experiences and the current policy can unexpectedly lead to a phenomenon called "soft divergence" in deep reinforcement learning training, where the agent's performance suffers\cite{vanhasselt2018deep}.

In this paper, we introduce a novel experience replay prioritization method Corrected Uniform Experience Replay (CUER) which balances the sampling of the transitions and makes the sampling distributions more similar to uniform if the entire transition history is considered.

Our main contributions are summarized as:
\begin{itemize}
    \item A novel experience replay prioritization method that aims for fairness across the entire transition history.
    \item Efficient and simple implementation that allows for substitution of any sampling method.
\end{itemize}
We performed our experiments and evaluated our results on OpenAI Gym and MuJoCo environments \cite{brockman2016openai,todorov_erez_tassa_2012} and significantly improved the performance compared to the existing sampling methods in almost every environment while achieving a remarkably faster rate of convergence.

\section{Related Work}
The concept of experience replay was first introduced by Lin \cite{lin_er} as a technique for accelerating the process of credit/blame propagation. This involves presenting previous experiences to a learning agent through a memory mechanism with limited capacity, where more recent experiences are given greater representation through an exponential weighting scheme. Meanwhile, this notion has undergone a transformation, wherein it has been conventional to employ a buffer of considerable magnitude, i.e. $10^6$, without placing significant emphasis on recent transitions. This approach introduced a uniform sampling strategy to revisit previously stored experiences, a technique that was groundbreaking in the context of the Deep Q-Network (DQN) \cite{mnih2013playing} and proved to be highly successful. However, the uniform sampling strategy does not account for the varying impacts of different experiences, limiting learning efficiency in complex environments. To address this, Prioritized Experience Replay (PER) \cite{schaul2016prioritized} is an extension of this concept, where they first introduced the notion of relative importance or priority among different experiences using the temporal difference (TD) error. During the training, the transitions are sampled considering these priorities, enabling the agent to learn significant experiences that might be ignored if a uniform sampling is used. It is an influential extension of the concept due to its success in a variety of tasks and its practical implementation.

Combined Experience Replay (CER) \cite{zhang2017deeper} is an extension of experience replay that can be treated as a plug-in to the standard mechanism. In a standard application of an experience replay, large buffer sizes are usually preferred to preserve the earlier transition. However, the stochastic nature of the replay buffer combined with a large buffer size creates another problem where the agent might not use valuable transitions at all, e.g., the most recent transition in an environment that responds better to an on-policy mechanism. CER solves this problem by combining the most recent transition with the sampled batch of transitions in favor of guaranteeing the usage of each experience at least once, reducing the negative impact of the large buffer size in the experience replay. Hindsight Experience Replay (HER) \cite{andrychowicz2017hindsight} is a separate extension of the standard experience replay that enables agents to learn complicated behaviors even in sparse and binary reward settings. Complex tasks, i.e., robotic manipulation, usually suffer from sparse and binary rewards, as it is difficult to assess continuous and informative extrinsic rewards to a learning agent through the trajectory without objective success criteria, which is usually presented as a goal state. HER overcomes this lack of reward by incorporating additional goals into each experience other than the actual goal state, even if it does not achieve the intended goal. This allows agents to learn from their failure experiences, increasing the sample efficiency, making learning possible in a sparse and binary reward setting, and removing the need for tailored rewards in such tasks. Both approaches are orthogonal to CUER and can be easily combined.

Experience Replay Optimization (ERO) \cite{zha2019experience} is the first study that integrates learning into an experience replay framework to maximize the cumulative rewards. ERO uses a priority score function, namely the replay policy, to determine the replay probability of each transition using empirically selected features and a deep neural network as a function approximator. This neural network is trained separately, where the learning process alternates between the agent policy and the replay policy. The priority score function evaluates each transition in the replay buffer and stores the results in an additional vector for later sampling. To further improve the sampling process, they adopted a Bernoulli distribution to avoid disregarding transitions with low scores. The ultimate collection of sampled experiences is the result of a two-stage sampling procedure, wherein a subset of experiences is chosen using the Bernoulli distribution and subsequently subjected to uniform sampling. The Neural Experience Replay Sampler (NERS) \cite{oh2021learning} improved replay learning by introducing the notion of relative importance between transitions. Instead of conducting individual evaluations for each transition like ERO, NERS additionally took into account the relative importance between experiences in the replay buffer by utilizing global features. Moreover, NERS increased the number of features to be considered to evaluate the transitions and used a sum-tree structure \cite{schaul2016prioritized} for sampling to further improve its effectiveness and efficiency. However, the focus of this study does not include the examination of replay learning methods. Instead, our research is limited to prioritized sampling approaches that can be seamlessly integrated without the need for further learning procedures.

\section{Background}
In this section, we explore the concept of reinforcement learning, with a focus on two key off-policy continuous control algorithms: Twin Delayed Deep Deterministic Policy Gradient (TD3) \cite{fujimoto2018addressing} and Soft Actor Critic (SAC) \cite{soft_actor_critic_paper}. We also examine various experience replay methods, emphasizing their innovations and performance implications. 

\subsection{Reinforcement Learning}
Reinforcement Learning (RL) involves an agent learning to maximize cumulative rewards thorugh interaction with an environment. This process is modeled as a Markov Decision Process (MDP), where at each discrete time step \(t \), the agent observes a state \( s_t \in S \), chooses an action $a_t \in A$ according to its policy $\pi(a|s_t)$, receives a reward, and transitions to the next state $s'_t$ based on the environment's dynamics $P(s',r|s,a)$. Each set of these elements -$(s,a,r,s')$- forms a transition, stored in a replay buffer for future learning. The agent aims to maximize its return, defined as the discounted sum of future rewards: $G_t = \sum_{i=t}^{T} \gamma^{i-t} r(s_i, a_i)$, where $\gamma$ is the discount factor. 

\subsection{Twin Delayed Deep Deterministic Policy Gradient (TD3)}

Twin Delayed Deep Deterministic Policy Gradient (TD3) is an advanced reinforcement learning algorithm that builds upon the Deep Deterministic Policy Gradient (DDPG) framework \cite{fujimoto2018addressing}. TD3 addresses the challenge of overestimation bias, a common issue in value-based methods, particularly in environments with high-dimensional and continuous action spaces \cite{fujimoto2018addressing, soft_actor_critic_paper}.

The key innovations of TD3 are its twin Q-function architecture and delayed policy updates, which together significantly reduce overestimation bias and improve learning stability. The algorithm maintains two separate Q-functions (critics), \( Q_{\theta_1} \) and \( Q_{\theta_2} \), and uses the minimum of their estimates for value updates, effectively mitigating the overestimation tendency of single Q-function approaches. The target value for updating these Q-functions is given by:

\[ y = r + \gamma \min_{i=1,2} Q_{\theta'_i}(s', \pi_{\phi'}(s')) \]

where \( \theta'_i \) are the parameters of the target networks and \( \pi_{\phi'} \) is the target policy network. This approach prevents the amplification of estimation errors that can occur in standard Q-learning updates.

Additionally, TD3 introduces policy update delays, meaning the policy (actor) is updated less frequently than the value functions. This delay helps in decoupling the policy and value updates, further stabilizing the training process.

TD3 also incorporates target policy smoothing, a technique that adds noise to the target policy's actions to smooth out Q-value estimates, preventing exploitation of Q-function errors by the policy.

These improvements enable TD3 to achieve more stable and reliable learning, particularly in complex and high-dimensional environments. It demonstrates superior performance compared to DDPG, especially in terms of sample efficiency and robustness against hyperparameter variations \cite{fujimoto2018addressing}. 

\subsection{Soft Actor-Critic (SAC)}

Soft Actor-Critic (SAC) is a state-of-the-art off-policy algorithm in reinforcement learning, particularly effective in continuous action spaces. SAC is distinguished by its incorporation of entropy into the reward framework, promoting exploration by maximizing a combination of expected return and entropy. This approach ensures a balance between exploring new actions and exploiting known rewards, leading to robust and effective policies \cite{soft_actor_critic_paper}.

The objective function of SAC is expressed as:

\[ J(\pi) = \mathbb{E}_{(s_t, a_t) \sim \rho_\pi} \left[ Q_{\theta}(s_t, a_t) - \alpha \log \pi(a_t|s_t) \right] \]

where \( \rho_\pi \) denotes the state-action distribution under policy \( \pi \), \( Q_{\theta} \) is the action-value function parameterized by \( \theta \), and \( \alpha \) represents the temperature parameter which controls the trade-off between exploration (entropy) and exploitation (reward). The term \( \log \pi(a_t|s_t) \) signifies the policy's entropy, encouraging the policy to be stochastic and hence, exploratory.

SAC employs a twin-critic setup similar to TD3, utilizing two Q-functions to mitigate positive bias in the policy improvement step inherent to function approximation errors. The policy is updated to maximize the minimum of these two Q-function estimates. Furthermore, SAC updates the temperature parameter \( \alpha \) automatically, adapting the degree of exploration to the specific requirements of the task.

SAC's performance in various benchmark tasks, particularly in high-dimensional and complex environments, demonstrates its superiority in terms of sample efficiency, robustness, and stability \cite{soft_actor_critic_paper}. This robustness is attributed to its entropy-based exploration, which prevents premature convergence to suboptimal deterministic policies and facilitates a more thorough exploration of the action space.

\subsection{Experience Replay Methods}
Experience Replay is a fundamental technique in reinforcement learning that involves storing and reusing past experiences of the agent\cite{lin_er}. Its first adaptation to deep reinforcement learning involved uniformly sampling transitions from the replay buffer \cite{mnih2013playing}. However, it was soon recognized that not all transitions contribute equally to the learning process. Some are more informative or 'surprising' to the agent, thereby offering greater learning potential. This realization led to the development of more sophisticated sampling methods.

One of the most notable advancements in the experience replay mechanism is Prioritized Experience Replay (PER). PER shifts the focus from uniform to prioritized sampling, increasing the probability of selecting transitions that yield unexpected outcomes for the agent \cite{schaul2016prioritized}. The 'unexpectedness' or importance of a transition is quantified using Temporal Difference (TD) Error as a proxy \cite{schaul2016prioritized}. The TD Error for a transition is defined as:
\[
\delta = \left| r + \gamma Q(s', a'; \theta') - Q(s, a; \theta) \right|
\]
where $\theta$ and $\theta'$ represent the parameters of the current and target value networks, respectively. The absolute value of this difference indicates how surprising or informative the transition is the the agent, under the assumption that a high TD error corresponds to a high degree of unexpectedness.
The sharpness of the prioritization in PER is modulated using parameter $\alpha$, which blends prioritized with uniform sampling. The sampling probability of a transition becomes: 
\[
P(i) = \frac{p_i^{\alpha}}{\sum_k p_k^{\alpha}}
\]
This formulation allows for a smooth adjustment between purely uniform and fully prioritized sampling, enabling a balance that can be tuned according to the specific needs of the learning task.

Beyond PER, other experience replay methods have been developed with specific goals in mind. Hindsight Experience Replay (HER) is particularly effective in environments with sparse and binary rewards. It tackles the challenge of learning from failures by reframing unsuccessful experiences as successful ones towards alternative goals. HER reinterprets each experience in the replay buffer in the context of multiple goals, not just the one originally intended. This re-interpretation significantly enriches the training data, allowing the agent to learn useful policies from what would otherwise be uninformative episodes. 

\section{Corrected Uniform Experience Replay (CUER)}

In this section, we discuss the problems that CUER aims to tackle and introduce the Experience Sampling Policy. Additionally, we provide a detailed explanation of how the experiences are selected from among potential candidates.

\subsection{Motivation}
In deep reinforcement learning, the conventional approach of uniformly sampling from an experience replay does not adequately acknowledge the varying significance of each transition. While uniform sampling offers simplicity and broad coverage, it often includes outdated transitions that may no longer be relevant, leading to a training process that strays from the optimal policy. On the other hand, increasing the sampling probability of specific transitions might intuitively seem like an improvement; however, this method risks overemphasizing certain experiences. Such heavy prioritization can lead to undesirable updates in both the Actor and Critic networks, with the extent of these detrimental effects often being proportional to the degree of prioritization. This creates a situation where transitions that are over-represented can push the training further off-policy, undermining the stability and effectiveness of the learning process.

Furthermore, the significance of transitions evolves throughout training as updates are made to the agent’s policy or the value network after each iteration. Consequently, to accurately adjust the importance of samples within the buffer, it would be necessary to traverse the entire replay buffer and recalculate the sampling probabilities. However, this becomes computationally impractical as the number of samples in the buffer grows quickly. For instance, PER uses Temporal Difference Error as the prioritization metric, and it recalculates the sampling probability of a transition only when it is sampled \cite{schaul2016prioritized}. Under this method, the expected interval between samplings of a transition can be given as follows:
\begin{equation}
    T_i = \frac{1}{P_i N} 
    \label{sampling_period_per}
\end{equation} 
where $P_i$ is the probability of sampling the $i$th transition at a timestep, and N is the batch size. This method presumes that the importance of a transition stays constant until it is sampled again. However, a transition initially deemed beneficial may become neutral or even detrimental as training progresses for the agent, and the reverse is also true \cite{zhang2017deeper}. Given these challenges, an algorithm that prioritizes transitions could result in a less effective training process. Consequently, the basic Experience Replay (ER) algorithm, which randomly samples transitions, might surpass a method that selectively prioritizes samples in the replay buffer \cite{zhang2017deeper}.

In reinforcement learning, the 'deadly triad' refers to the interplay of function approximation, bootstrapping, and off-policy learning. Algorithms that embody these three elements can experience unbounded value estimates, which can hinder the agent's learning progress \cite{vanhasselt2018deep}. Of these elements, function approximation is crucial, particularly when the state and action spaces of the task are vast, making it impractical to explore every state-action pair, especially in continuous domains. An alternative could be to adopt Monte Carlo learning, which avoids bootstrapping. However, Monte Carlo methods depend on complete trajectories that conclude at a terminal state, rendering them unsuitable for tasks without clear endpoints. Choosing On-Policy learning over Off-Policy prevents the agent from benefiting from experiences gathered under previous policies, which could lead to highly correlated transitions that disrupt neural network training \cite{vanhasselt2018deep}. The last aspect of the deadly triad, Off-Policy learning, can be mitigated by adjusting the sampling probabilities of transitions, thereby enhancing the algorithm's performance on learning tasks \cite{vanhasselt2018deep}.

Therefore, there exists a critical need for a refined sampling strategy that balances the representation of valuable transitions without disproportionately influencing the learning trajectory. In this paper, we present Corrected Uniform Experience Replay (CUER), a novel experience replay prioritization algorithm designed to minimize off-policy updates while training off-policy deep reinforcement learning algorithms. This method strategically samples stored experiences, ensuring fairness among all entries and adapting to the changing significance of transitions by making the sampled state distribution more closely aligned with the on-policy.

\subsection{Proposed Strategy: Dynamic Transition Priority Adjustment}

The proposed strategy involves initially assigning a high priority to new transitions added to the replay buffer. The priority of each transition is represented by its sampling probability, which is adjusted dynamically throughout the training process. The initial high priority ensures that new transitions are considered more frequently in the early stages after their addition.
\begin{itemize}
    \item \textbf{Initialization}:  For a new transition \( t_i \), set the initial priority probability \( P(t_i) \) as:
    \[
    P(t_i) = \frac{\text{batch\_size}}{\Psi}
    \]
    where \( \Psi \) is the total of sample priorities of transitions in the buffer.
\end{itemize}

\begin{itemize}
    \item \textbf{Priority Update upon Sampling}: Each time a transition is sampled, its priority is decreased to gradually reduce its sampling probability, promoting a fair chance for all transitions over time. The updated priority \( P'(t_i) \) after sampling is given by:
    \[
    P'(t_i) = \frac{P(t_i) * \Psi - 1}{\Psi}
    \]
    This decrement strategy prevents the dominance of any single transition in the sampling process, ensuring a more uniform exploration of the experience buffer.
\end{itemize}

\begin{itemize}
    \item \textbf{Sampling Probability Adjustment}: The sampling probability of each transition is recalculated after every training iteration to reflect the updated priorities, ensuring that the probability distribution across the transitions adjusts to the dynamic learning environment. The probability \( \text{Pr}(t_i) \) of sampling transition \( t_i \) is:
    \[
    \text{Pr}(t_i) = \frac{P(t_i)}{\sum_{j=1}^{N} P(t_j)}
    \]
    This normalization ensures that the sum of probabilities over all transitions remains equal to 1, maintaining a proper probability distribution.
\end{itemize}

CUER's Dynamic Priority Adjustment Strategy comes with its own advantages which are:
\begin{itemize}
    \item \textbf{Fairness}: By continuously adjusting the priorities based on sampling occurrences, the strategy ensures that all transitions have a fair chance of being selected over the course of training, preventing any bias towards older or newer transitions.
\end{itemize}

\begin{itemize}
    \item \textbf{Adaptability}: The strategy adapts to the evolving significance of transitions as the agent's policy updates, making the replay process more aligned with the current policy and enhancing the overall efficiency of learning.
\end{itemize}

\begin{itemize}
    \item \textbf{Simplicity and Efficiency}: The implementation of this priority adjustment is straightforward and does not require complex recalculations across the entire buffer, thus preserving computational resources and simplifying integration into existing systems. For an efficient implementation, sum-trees are used to assign priorities to the stored transitions dynamically.
\end{itemize}

This dynamic prioritization approach fosters a more balanced exploration of the experience space creating a close to uniform sampled experience space when whole transition history is considered, potentially leading to more stable and effective learning outcomes in deep reinforcement learning tasks.

\begin{figure*}[t]
\centering
%Objects need to be carried from kitchen counter (grey) to kitchen counter (grey)
\subfigure[Ant-v4 environment.]{
\includegraphics[width=0.30\textwidth]{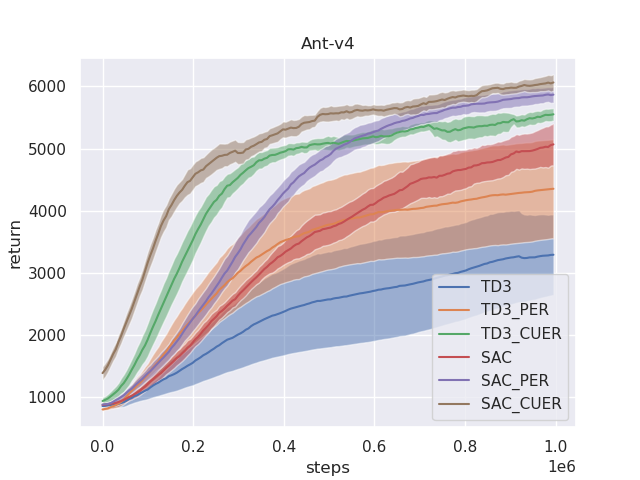}}
\subfigure[Hopper-v4 environment.]{
\includegraphics[width=0.30\textwidth]{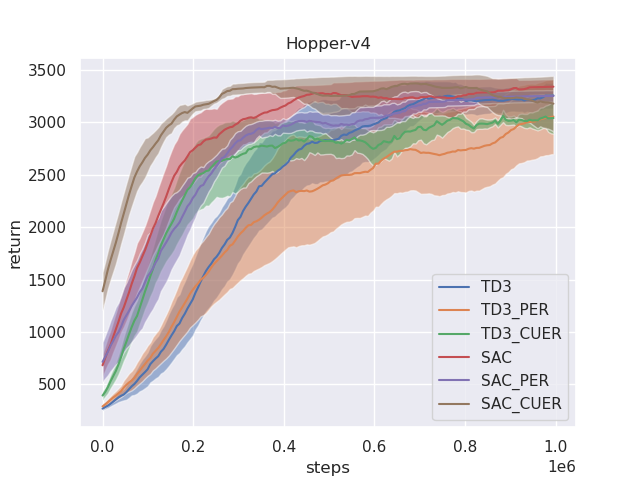}}
\subfigure[Humanoid-v4 environment.]{
\includegraphics[width=0.30\textwidth]{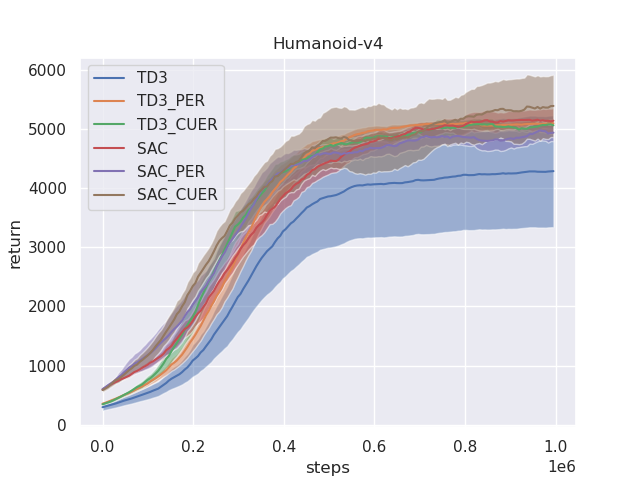}}
\subfigure[Walker2d-v4 environment.]{
\includegraphics[width=0.30\textwidth]{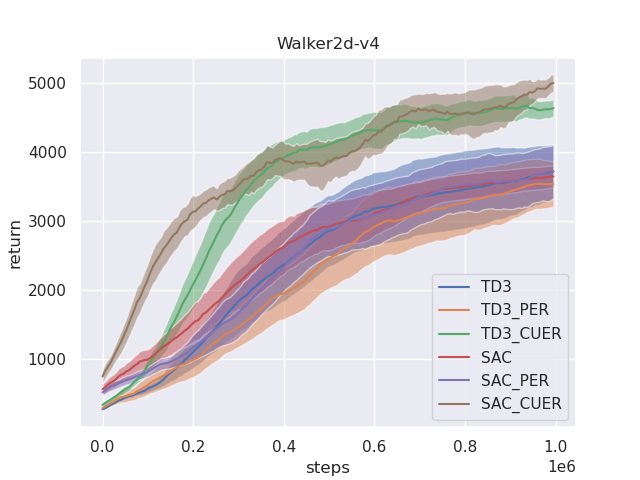}}
\subfigure[HalfCheetah-v4 environment.]{
\includegraphics[width=0.30\textwidth]{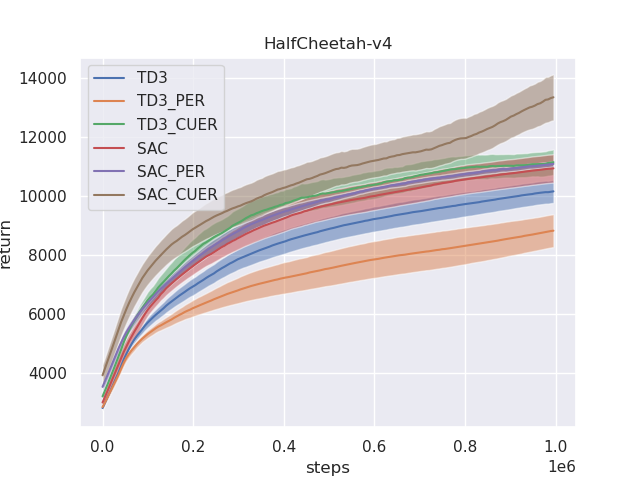}}
\subfigure[LunarLanderContinuous-v2 environment.]{
\includegraphics[width=0.30\textwidth]{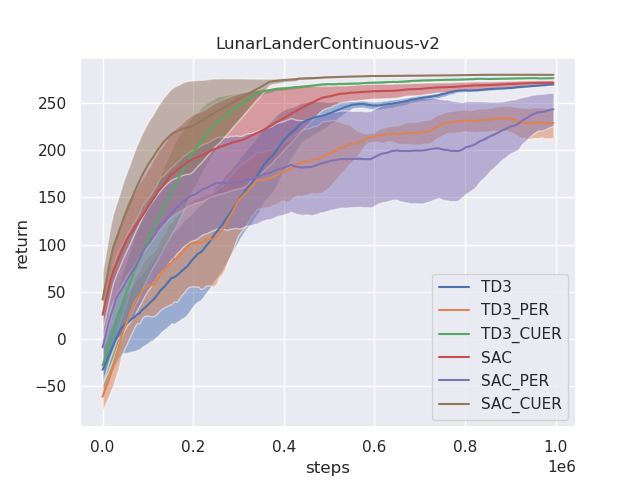}}
\caption{Comparison of CUER with SOTA baselines in various environments.}
\label{benchmark_results}
\end{figure*}

\section{Experiments}
We present Corrected Uniform Experience Replay (CUER), an experience replay method designed to ensure fairness during sampling across all transitions stored within the replay buffer. Our method is extensively evaluated in various continuous control tasks and compared with uniform sampling\cite{mnih2013playing}, Prioritized Experience Replay (PER)\cite{schaul2016prioritized}, and Corrected Experience Replay (CER)\cite{zhang2017deeper} using TD3\cite{fujimoto2018addressing} and SAC\cite{soft_actor_critic_paper} algorithms. Additionally, we investigate the impact of varying experience replay buffer sizes on uniform sampling to demonstrate the response of the tasks to recent transitions.
\subsection{Task Selection}
\label{subsection:task}

To evaluate our experience replay prioritization method, we measure its performance and that of existing approaches on various MuJoCo environments \cite{todorov_erez_tassa_2012} available through the Gym interface \cite{brockman2016openai, towers_gymnasium_2023}. Featured tasks are Ant-v4, HalfCheetah-v4, Hopper-v4, Humanoid-v4, LunarLanderContinous-v2, and Walker2d-v4. These environments serve as benchmark environments, enabling us to make a fair comparison with existing methods.
\begin{figure*}[!ht]
\centering
%Objects need to be carried from kitchen counter (grey) to kitchen counter (grey)
\subfigure[Ant-v4.]{
\includegraphics[width=0.2\textwidth]{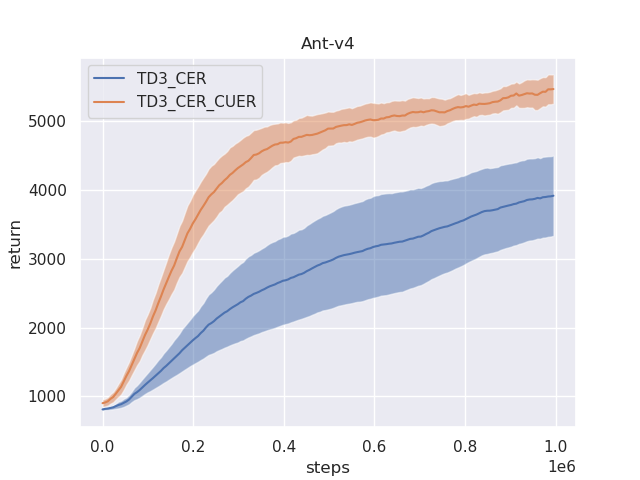}}
\subfigure[HalfCheetah-v4.]{
\includegraphics[width=0.2\textwidth]{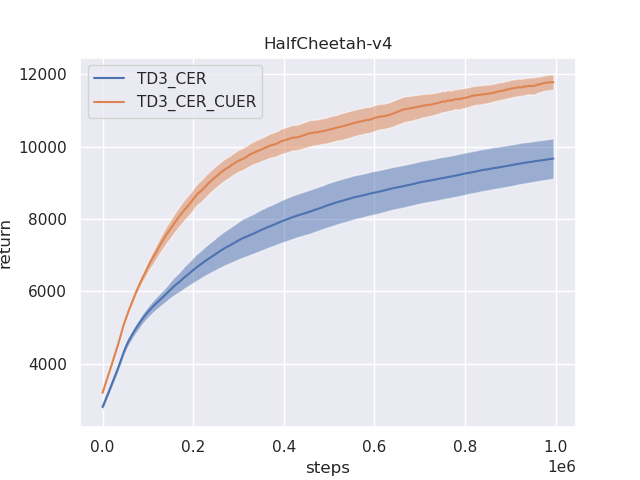}}
\subfigure[Hopper-v4.]{
\includegraphics[width=0.2\textwidth]{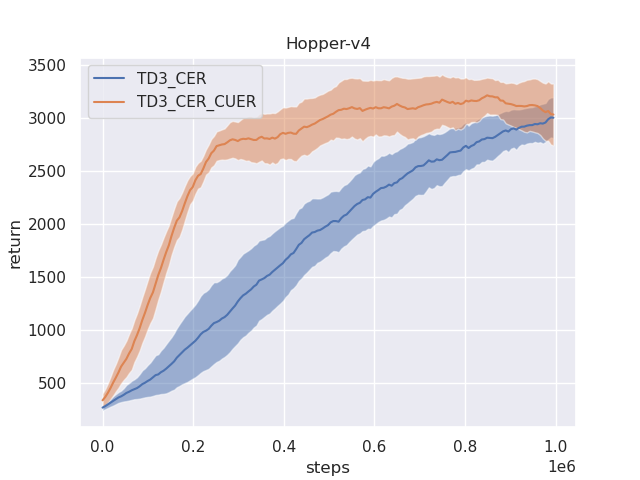}}
\subfigure[Humanoid-v4.]{
\includegraphics[width=0.2\textwidth]{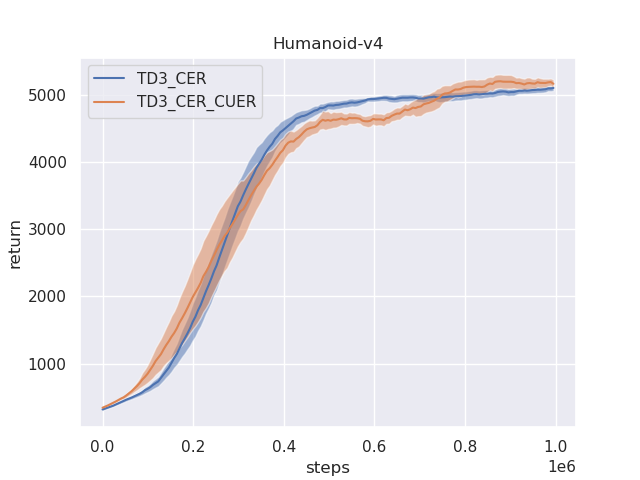}}
\caption{Comparison of TD3\_CER and TD3\_CER\_CUER in various environments.}
\label{cer_results}
\end{figure*}

\subsection{Benchmark Results}
In this section, we present the benchmark results of our proposed CUER algorithm in various environments mentioned in Section \ref{subsection:task}. The performance of CUER is compared against state-of-the-art (SOTA) baselines, including TD3, TD3 with Prioritized Experience Replay (TD3\_PER), SAC, and SAC with Prioritized Experience Replay (SAC\_PER).

As depicted in Figure \ref{benchmark_results}, the CUER algorithm consistently outperforms or matches the performance of existing SOTA baselines across almost all environments. Notably, CUER demonstrates significant improvements in convergence speed and reduction in variance. For instance, in the HalfCheetah-v4 environment, CUER improves the performance of the SAC agent, achieving results comparable to existing baselines in just 1 million training steps, which typically require 3 million steps.

Another important contribution is the significant impact of CUER on the TD3 algorithm. The results indicate that the second-best performance is usually achieved by the TD3 algorithm with CUER sampling. This highlights the effectiveness of our experience replay prioritization algorithm, demonstrating that CUER can significantly enhance the learning efficiency and stability of reinforcement learning agents.

Overall, the results validate the robustness and efficiency of the CUER algorithm, making it a valuable addition to the suite of techniques for improving reinforcement learning performance.
\subsection{Comparison with CER}

We present an additional section to compare our CUER algorithm with CER. CER ensures that the latest transitions, which include the recent policy's effect, are sampled more frequently. This approach reduces the off-policy component of the training, mitigating the risk of divergence, as discussed in the "deadly triad"\cite{vanhasselt2018deep}. Our approach shares a similar motivation. By aiming to create a uniform sampling distribution across the entire transition history, we prioritize recent transitions, ensuring the sampled distribution is less affected by older transitions.

CER employs uniform sampling beyond including the latest transitions, making it orthogonal to our approach. This characteristic makes it an excellent case study to demonstrate how CUER can enhance existing experience prioritization methods without disrupting their unique contributions. Our results indicate that combining CER with CUER outperforms the sole CER approach in all environments, while preserving the low variance and fast convergence properties.

As shown in Figure \ref{cer_results}, the combination of CER and CUER consistently outperforms the sole CER approach across all evaluated environments. This improvement demonstrates the effectiveness of CUER in enhancing existing experience prioritization methods, maintaining low variance, and achieving faster convergence.

\subsection{Investigation of Different Buffer Sizes}

It is plausible to question the difference between giving high priority to the latest transitions and decreasing the buffer size of the experience replay buffer to store more on-policy transitions. To address this, we conducted additional experiments comparing CUER with uniform sampling using buffer sizes of 100,000 and 250,000.

\begin{figure*}[!ht]
\centering
%Objects need to be carried from kitchen counter (grey) to kitchen counter (grey)
\subfigure[Ant-v4 environment.]{
\includegraphics[width=0.30\textwidth]{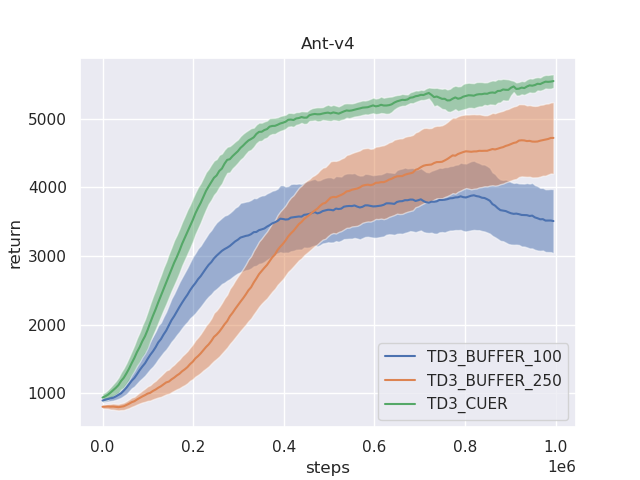}}
\subfigure[Hopper-v4 environment.]{
\includegraphics[width=0.30\textwidth]{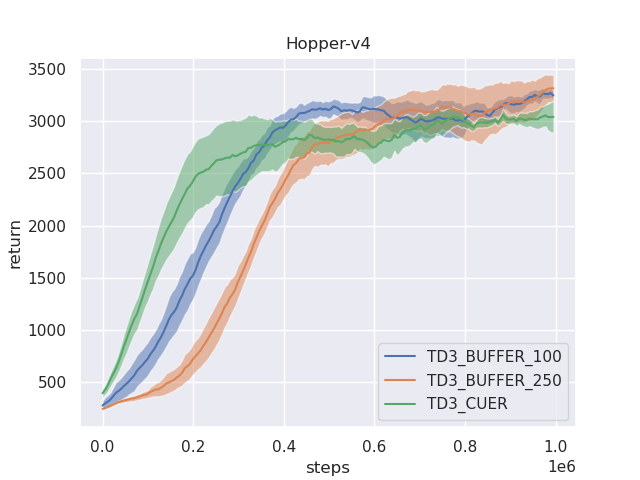}}
\subfigure[Humanoid-v4 environment.]{
\includegraphics[width=0.30\textwidth]{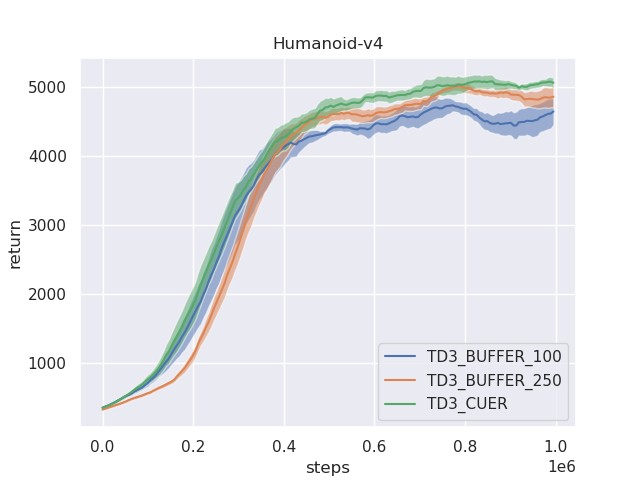}}
\subfigure[Walker2d-v4 environment.]{
\includegraphics[width=0.30\textwidth]{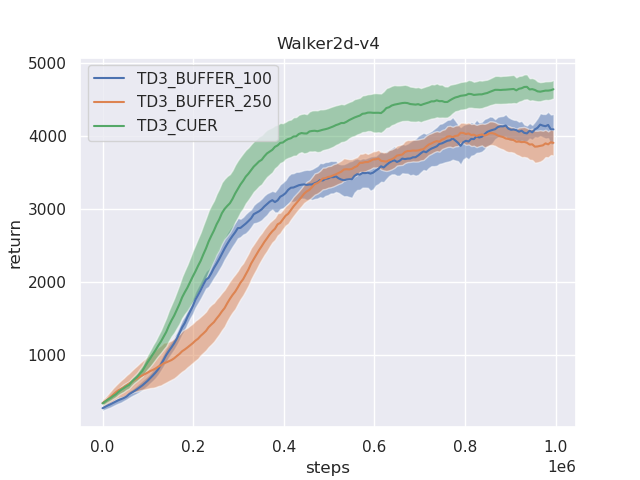}}
\subfigure[HalfCheetah-v4 environment.]{
\includegraphics[width=0.30\textwidth]{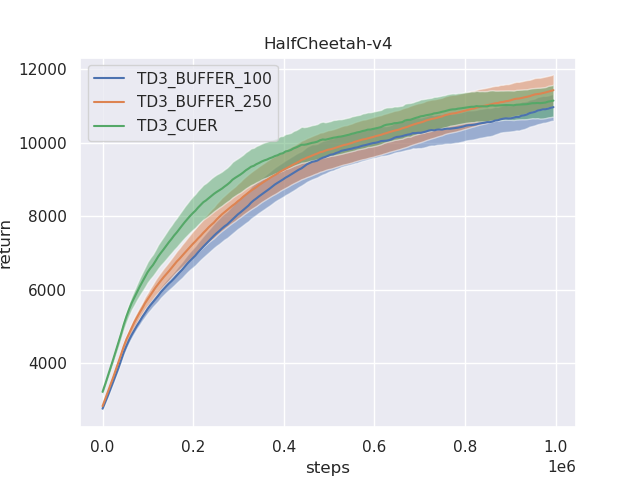}}
\subfigure[LunarLanderContinuous-v2 environment.]{
\includegraphics[width=0.30\textwidth]{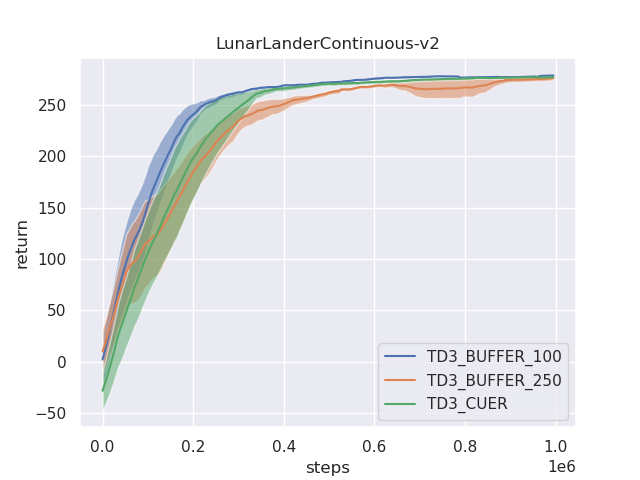}}
\caption{Comparison of of CUER with Uniform Sampling having different buffer sizes in various environments.}
\label{buffer_results}
\end{figure*}

The results, depicted in Figure \ref{buffer_results}, support that our approach is distinct from simply decreasing the buffer size. Although we prioritize the latest transitions, the buffer still stores older transitions, allowing them to be sampled stochastically. This process ensures the sampling distribution covers the entire transition history rather than a limited portion.

CUER consistently outperforms or achieves comparable results to the baselines, even with reduced buffer sizes. It is also noteworthy that the baselines demonstrate stable behavior, as the transitions stored in the buffer are closer to the policy. However, CUER still converges faster than its competitors, proving it to be an effective experience replay prioritization method.

\section{Conclusion}

In this paper, we presented Corrected Uniform Experience Replay (CUER), a novel experience replay prioritization method designed to enhance the performance of off-policy continuous control algorithms. CUER addresses the inherent biases and inefficiencies of conventional uniform sampling by dynamically adjusting the sampling probabilities of transitions, ensuring a more balanced and fair representation of experiences. Our approach prioritizes recent transitions while maintaining the ability to sample from the entire transition history, thus mitigating the negative impacts of off-policy updates and enhancing learning stability.

Through extensive experiments in various MuJoCo environments, CUER demonstrated significant improvements in convergence speed, variance reduction, and final performance when compared to state-of-the-art baselines, including TD3, TD3 with Prioritized Experience Replay (TD3\_PER), SAC, and SAC with Prioritized Experience Replay (SAC\_PER). Moreover, CUER showed remarkable performance gains when combined with CER, further validating its effectiveness as an experience replay prioritization method.

Our investigations also highlighted that CUER's advantages are distinct from merely decreasing the experience replay buffer size, as it preserves the ability to sample older transitions stochastically, ensuring comprehensive coverage of the transition history. This makes CUER a robust and efficient solution for reinforcement learning tasks, contributing to the advancement of experience replay techniques.

Overall, CUER offers a promising approach to improving the learning efficiency and stability of off-policy deep reinforcement learning algorithms, making it a valuable addition to the toolkit of reinforcement learning practitioners.

\bibliography{ref}
\bibliographystyle{ieeetr}

\end{document}